\documentclass[10pt,journal,compsoc]{IEEEtran}
\usepackage[nocompress]{cite}
\usepackage[pdftex]{graphicx}
\usepackage{amsmath}
\usepackage{algorithmic}
\usepackage{array}
\usepackage[caption=false,font=footnotesize,labelfont=sf,textfont=sf]{subfig}
\usepackage{fixltx2e}
\usepackage{stfloats}
\ifCLASSOPTIONcaptionsoff
  \usepackage[nomarkers]{endfloat}
 \let\MYoriglatexcaption\caption
 \renewcommand{\caption}[2][\relax]{\MYoriglatexcaption[#2]{#2}}
\fi
\usepackage{url}

\usepackage{booktabs}
\usepackage{amssymb}
\usepackage{makecell}
\usepackage{multirow}
\usepackage{bm}
\usepackage{color}
\usepackage{comment}
\usepackage{threeparttable}
\usepackage{pdfpages}

\newcommand{\eg}{e.g.}
\newcommand{\ie}{i.e.}

\newcommand{\etal}{et al.}

\newtheorem{definition}{Definition}[section]
\newtheorem{theorem}{Theorem}[section]

\begin{document}
\title{Adversarial Margin Maximization Networks}

\author{Ziang Yan, Yiwen Guo,
 and Changshui Zhang,~\IEEEmembership{Fellow,~IEEE}
 \IEEEcompsocitemizethanks{\IEEEcompsocthanksitem Z. Yan and C. Zhang are with the Institute for Artificial Intelligence, Tsinghua University (THUAI), the State Key Lab of Intelligent Technologies and Systems, the Beijing National Research Center for Information Science and Technology (BNRist), the Department of Automation, Tsinghua University, Beijing 100084, China. E-mail: yza18@mails.tsinghua.edu.cn; zcs@mail.tsinghua.edu.cn.
  \IEEEcompsocthanksitem Y. Guo is with the Bytedance AI Lab, Beijing, China. E-mail: guoyiwen.ai@bytedance.com.}
 \thanks{}}



\IEEEtitleabstractindextext{
 \begin{abstract}
  The tremendous recent success of deep neural networks (DNNs) has sparked a surge of interest in understanding their predictive ability.
  Unlike the human visual system which is able to generalize robustly and learn with little supervision, DNNs normally require a massive amount of data to learn new concepts.
  In addition, research works also show that DNNs are vulnerable to adversarial examples---maliciously generated images which seem perceptually similar to the natural ones but are actually formed to fool learning models, which means the models have problem generalizing to unseen data with certain type of distortions.
  In this paper, we analyze the generalization ability of DNNs comprehensively and attempt to improve it from a geometric point of view.
  We propose adversarial margin maximization (AMM), a learning-based regularization which exploits an adversarial perturbation as a proxy.
  It encourages a large margin in the input space, just like the support vector machines.
  With a differentiable formulation of the perturbation, we train the regularized DNNs simply through back-propagation in an end-to-end manner.
  Experimental results on various datasets (including MNIST, CIFAR-10/100, SVHN and ImageNet) and different DNN architectures demonstrate the superiority of our method over previous state-of-the-arts.
  Code and models for reproducing our results will be made publicly available.
 \end{abstract}

 }

\maketitle
\IEEEdisplaynontitleabstractindextext
\IEEEpeerreviewmaketitle

\IEEEraisesectionheading{\section{Introduction}\label{sec:intro}}
\IEEEPARstart{A}lt{hough} deep neural networks (DNNs) have shown state-of-the-art performance on a variety of challenging computer vision tasks, most of them are still "notorious" for requiring massive amount of training data.
In addition, a bunch of recent works also demonstrate that DNNs are vulnerable to adversarial examples~\cite{Goodfellow2015, Nguyen2015, Moosavi2016, Papernot2018}, indicating the models have problem generalizing to unseen data with possible types of distortions~\cite{Szegedy2014, Miyato2017}.
These undesirable facts motivate us to analyze the generalization ability of DNNs and further find a principled way to improve it.

Our work in this paper stems specifically from a geometric point of view.
We delve deeply into the margin of DNNs and advocate a large margin principle for training networks.
Generally, such a principle can enhance the obtained models over two aspects:
(a). maximum margin classifiers usually possess better generalization ability, which is both theoretically guaranteed and empirically verified~\cite{Cortes1995, Vapnik1999, Platt2000, Xu2012};
and (b). they can also be naturally more robust to adversarial examples since it takes more efforts to perturb the inputs towards decision boundaries. 
This paper substantially extends the work of Yan \etal~published at NeurIPS 2018~\cite{Yan2018}.

The original concept of the large margin principle dates back to last century.
Novikoff, Cortes and Vapnik proved essential theorems for the perceptron~\cite{Novikoff1962} and support vector machines (SVMs)~\cite{Cortes1995}, based on a geometric margin $\gamma$ and an assumption that the training data can be separated with it:
\begin{equation}
 \sup_w\min_i y_i f(\mathbf x_i, w)>\gamma.
\end{equation}

Unlike the well-known scheme in linear classification~\cite{Cortes1995}, the geometric margin of nonlinear DNNs scarcely has close-form solutions, making it non-trivial to get incorporated in the training objective.
Although many attempts have been made to pursue this target, most of them just focus on a margin in the representation space (a.k.a, feature space)~\cite{Tang2013, Hu2014, Liu2016, Wang2018}.
Learned representations in such a manner may show better intra-class compactness and inter-class discriminability, but in practice a large margin in the feature space itself does not necessarily guarantee a large margin in the input space~\cite{An2015}, on account of the distance distortions of nonlinear hierarchical models.
To address this problem, several recent methods have been developed to suggest a large margin directly in the input space.
For example, An \etal~\cite{An2015} study contractive mappings in DNNs and propose contractive rectifier networks, while Sokoli\'{c} \etal~\cite{Sokolic2017} try penalizing the Frobenius norm of a Jacobian matrix instead.
These methods show significantly superior performance in scenarios where for instance the amount of training data is severely limited.
However, aggressive assumptions and approximations seem inevitable in their implementations, making them less effective in practical scenarios where the assumptions are not really satisfied.

In this paper, we propose adversarial margin maximization (AMM), a learning-based regularization that exploits an adversarial perturbation as a proxy of $\gamma$.
Our core idea is to incorporate an $l_2$ norm-based adversarial attack into the training process, and leverage its perturbation magnitude as an estimation of the geometric margin.
Current state-of-the-art attacks typically achieve $\sim$100\% success rate on powerful DNNs~\cite{Moosavi2016, CW2017, Dong2018}, while the norm of perturbation can be reasonably small and thus fairly close to the real margin values.
Since the adversarial perturbation is also parameterized by the network parameters (including weights and biases), our AMM regularizer can be jointly learned with the original objective through back-propagation.  
We conduct extensive experiments on MNIST, CIFAR-10/100, SVHN and ImageNet datasets to testify the effectiveness of our method.
The results demonstrate that our AMM significantly improves the test-set accuracy of a variety of DNN architectures, indicating an enhanced generalization ability as the training-set accuracies remain similar.

The rest of this paper is organized as follows.
In Section~\ref{sec:related}, we introduce representative margin-based methods for training DNNs.
In Section~\ref{sec:method}, we highlight our motivation and describe our AMM in detail.
In Section~\ref{sec:exp} we experimentally validate the effectiveness of our proposed method.
Finally, Section~\ref{sec:conclusion} draws the conclusions.

\section{Related Work}\label{sec:related}
Margin-based methods have a long research history in the field of pattern recognition and machine learning.
Theoretical relationships between the generalization ability and geometric margin of linear classifiers have been comprehensively studied~\cite{Vapnik1999} and the idea of leveraging large margin principle and constructing maximal margin separating hyperplane~\cite{Vapnik1999} also act as essential ingredients of many classical learning machines, c.f. the famous SVM~\cite{Cortes1995}.

Benefit from its solid theoretical foundation and intuitive geometric explanations, the large margin principle has been widely applied to a variety of real-world applications such as face detection~\cite{Osuna1999}, text classification~\cite{Tong2001}, gene selection for cancer classification~\cite{Guyon2002}, etc.
Nevertheless, there is as of yet few methods for applying such principle to DNNs which are ubiquitous tools for solving modern machine learning and pattern recognition tasks (but are also structurally very complex and generally considered as black-boxes).
This is mostly because the margin cannot be calculated analytically in general as with SVM.

In view of such opportunities and challenges, one line of researches targets at improving a "margin" in the representation space of DNNs instead.
For instance, Tang~\cite{Tang2013} replaces the final softmax layer with a linear SVM to maximize a margin in the last layer.
Hu \etal~\cite{Hu2014} propose a discriminative learning method for deep face verification, by enforcing the distance between each positive pair in the representation space to be smaller than a fixed threshold, and that to be larger than another threshold for all negative pairs, where a margin is naturally formed.
A similar strategy, first invented by Weinberger \etal~\cite{Weinberger2009} and dubbed as triplet loss, has also been widely adopted to many face recognition systems, \eg, FaceNet~\cite{Schroff2015}.
Sun \etal~\cite{Sun2016} theoretically study a margin in the output layer of DNNs, and propose a way of reducing empirical margin errors.
In the same spirit, Wang \etal~\cite{Wang2018} propose to further enhance the discriminability of DNN features via an ensemble strategy.

Stick with the representation space, some recent works also advocate large margins under different ``metrics", \eg, cosine similarity~\cite{Wang2018cosface} and the angular distance between logit vectors and the ground-truth~\cite{Liu2016 ,Liu2017, Deng2018}.
In essence, these methods maximize the inter-class variance while minimizing the intra-class variance, and thus the learned representations can be more discriminative.
However, as previously discussed~\cite{An2015}, owing to the high structural complexity of DNNs and possible distance distortions of nonlinear models, a large margin in the feature space does not necessarily assure a large margin in the input space.
That being said, the aforementioned benefits in Section~\ref{sec:intro} are not guaranteed.
See section~\ref{sec:exp} for some empirical analyses.
It is also worthy of mentioning that some previous works suggest that DNNs trained using stochastic gradients converge to large margin classifiers, but the convergence speed is very slow~\cite{Soudry2018, Wei2018}.

A few attempts have also been made towards enlarging the margin in input spaces.
In a recent work, An \etal~\cite{An2015} propose contractive rectifier networks, in which the input margin is proved to be bounded from below by an output margin which is inherently easier to optimize by leveraging off-the-shelf methods like SVM.
Sokoli\'{c} \etal~\cite{Sokolic2017} reveal connections between the spectral norm of a Jacobian matrix and margin, and try to regularize a simplified version of its Frobenius norm.
Contemporaneous with our work, Elsayed \etal~\cite{Elsayed2018} propose to use a one-step linear approximation to DNN mappings and enlarge the margin explicitly.
These methods are closely related to ours, but their implementations require rough approximations and can be suboptimal in many practical applications. Some detailed discussions are deferred to Section~\ref{sec:method_discussion} and experimental results for comparing with them will be provided in Section~\ref{sec:exp}. Ding \etal~\cite{Ding2018} also aim to approximate the margin more accurately but their method differs with ours in multiple ways as will be discussed in the Appendix.


\section{Adversarial Margin Maximization}\label{sec:method}

In this section, we introduce our method for pursuing large margin in DNNs.
First, we briefly review our recent work of deep defense~\cite{Yan2018} aimed at training DNNs with improved adversarial robustness.
We believe its functionality can be naturally regarded as maximizing the margin.
We then formalize the definition of margin and provide discussions of the generalization ability based on it.
Finally, we introduce our AMM for improving the generalization ability of DNNs.

\subsection{Our Deep Defense}\label{sec:dd}

Deep Defense improves the adversarial robustness of DNNs by endowing models with the ability of learning from attacks~\cite{Yan2018}.  
On account of the high success rate and reasonable computation complexity, it chooses DeepFool~\cite{Moosavi2016} as a backbone and tries to enlarge the $l_p$ norm of its perturbations.~\footnote{There exist stronger attacks which approximate the margin more precisely (like the Carlini and Wagner's~\cite{CW2017}), but they are computationally more complex; we have demonstrated that defending DeepFool helps to defend Carlini and Wagner's attack~\cite{Yan2018} as well therefore it can be a reasonable proxy.}

Suppose a binary classifier $f:\mathbb R^m \rightarrow \mathbb R$, where the input $\mathbf{x}$ is an $m$-dimensional vector and the predictions are made by taking the sign of classifier's outputs.
DeepFool generates the perturbation $\Delta_{\mathbf{x}}$ with an iterative procedure.
At the $i$-th step ($0\leq i < u$), the perturbation $\mathbf r^{(i)}$ is obtained by applying first-order approximation to $f$ and solving:
\begin{equation}\label{eq:minr}
 \min_{\mathbf r} \| \mathbf r \|_p \quad \mathrm{s.t.}\, f (\mathbf x + \Delta_\mathbf{x}^{(i)}) +  \nabla f (\mathbf x + \Delta_\mathbf{x}^{(i)})^T \mathbf r=0,
\end{equation}
where $\nabla f$ denotes $\frac{\partial f(\mathbf{x})}{\partial\mathbf x}$, and $\Delta_\mathbf{x}^{(i)} := \sum _{j=0}^{i-1} \mathbf r^{(j)}$.
Problem \eqref{eq:minr} can be solved analytically:
\begin{equation}\label{eq:df_single_step}
 \mathbf r^{(i)} = -\frac{f (\mathbf x + \Delta_\mathbf{x}^{(i)})}{\|\nabla f(\mathbf x + \Delta_\mathbf{x}^{(i)})\|_2^2} \nabla f (\mathbf x + \Delta_x^{(i)}).
\end{equation}
After computing all the $\mathbf r^{(i)}$s sequentially, the final DeepFool perturbation $\Delta_{\mathbf{x}}$ is obtained by adding up the $\mathbf r^{(i)}$s obtained from each step:
\begin{equation}
 \Delta_{\mathbf x}= \mathbf r^{(0)}+...+\mathbf r^{(u-1)},
\end{equation}
where $u$ is the maximum iteration allowed.
If the prediction class of $f$ changes at any iteration before the $u$-th, the loop should terminate in advance.
Such procedure directly generalizes to multi-class cases, as long as a target label is properly chosen at each iteration.
Given a baseline network, the procedure may take about 1-3 iterations to converge on small datasets such as MNIST/CIFAR-10, and 3-6 iterations on large datasets like ImageNet~\cite{Deng2009}\footnote{For example, it convergences within 3 iterations on all the MNIST images with a 5-layer LeNet reference model, and 6 iterations on 99.63\% of the ImageNet images with a ResNet-18 reference model.}.

\subsubsection{Regularization and High Order Gradients}
In fact, the $\ell_p$ norm of $\Delta_{\mathbf x}$ is a popular metric for evaluating the adversarial attacks and the robustness of DNNs~\cite{Moosavi2016, CW2017, Cisse2017}.
Given an input vector $\mathbf{x}$, the gradient $\nabla f(\mathbf{x})$, as well as the perturbation $\Delta_{\mathbf x}$, are both parameterized by the learnable parameters of $f$.
Consequently, in order to give preference to those $f$ functions with stronger robustness, one can simply penalize the norm of $\Delta_{\mathbf x}$ as a regularization term during training.
With modern deep learning frameworks such as PyTorch~\cite{Pytorch2017} and TensorFlow~\cite{Tensorflow2015}, differentiating $\Delta_{\mathbf x}$ can be done via automatic derivation with higher order gradients.
One might also achieve this by building a ``reverse'' network to mimic the backward process of $f$, as described in~\cite{Yan2018}.

We emphasize the high order gradients form an essential component in our method and.
In principle, if no gradient flows through $\nabla f$, the regularization can be viewed as maximizing the norm in a normalized logit space, given $\Vert\nabla f \Vert$ as the normalizer.
Considering possible distance distortions of nonlinear hierarchical DNNs, it is definitely less effective in influencing the perturbation or whatever else.
Experimental analysis will be further given in Section~\ref{sec:exp_ho} to verify the importance of the high order gradients.

\subsubsection{Correctly and Incorrectly Classified Samples}
Deep Defense applies different regularization strategies on correctly and incorrectly classified samples.
Specifically, if an input is correctly classified during training, we expect it to be pushed further away from the decision boundary, thus a smaller value of $\Vert\Delta_{\mathbf x}\rVert_p$ is anticipated.
In practice, the target class at each iteration is chosen to be the one results in the smallest $\mathbf r^{(i)}$.
Conversely, if an input is misclassified by the current model, we instead expect it to be closer to the decision boundary (between its current prediction and the ground-truth label), since we may intuitively hope the input sample to be correctly classified by the model in the future.
The target class is always set to be the ground-truth and a larger value of $\Vert\Delta_{\mathbf x}\rVert_p$ is anticipated.

In summary, the Deep Defense regularizer can be written as:
\begin{equation}\label{eq:regularizer0}
 L_{\mathrm{DD}}=\sum_{k\in \mathcal T} R\left(-c\frac{\|\Delta_{\mathbf x_k} \|_p}{\|\mathbf x_k\|_p}\right) +\sum_{k\in \mathcal F} R\left(d\frac{\|\Delta_{\mathbf x_k} \|_p}{\|\mathbf x_k\|_p}\right),
\end{equation}
or similarly
\begin{equation}\label{eq:regularizer1}
 L_{\mathrm{DD}}=\frac{1}{n}\sum_{k\in \mathcal T} R\left(-c\|\Delta_{\mathbf x_k} \|_p\right) +\frac{1}{n}\sum_{k\in \mathcal F} R\left(d\|\Delta_{\mathbf x_k} \|_p\right),
\end{equation}
where $n$ is the number of training samples, $\mathcal T$ is the index set of correctly classified training examples, $\mathcal F$ is its complement, $c,d>0$ are two hyper-parameters balancing these two groups of samples, $R$ is the shrinkage function that balances examples within the same group (details in Section \ref{sec:large_margin_dnns}).
The sets $\mathcal T$ and $\mathcal F$ are updated in each training iteration.
The whole optimization problem is given by:
\begin{equation}\label{eq:whole_opt_problem}
 \min_{\mathcal W}\, \sum_k L(\mathbf y_k, f(\mathbf x_k; \mathcal W)) + \lambda L_{\mathrm{DD}} + \alpha L_{\mathrm WD},
\end{equation}
where $L$ is the original classification objective (\eg, cross-entropy or hinge loss), $\lambda$ is the coefficient for regularizer, and $\alpha L_{\mathrm WD}$ is the weight decay term.
We adopt the unnormalized version~\eqref{eq:regularizer1} in this paper, since it connects the most to the margin to be defined in Section~\ref{sec:lims}, and further the generalization ability.

\subsection{Margin, Robustness and Generalization}\label{sec:lims}

Deep defense achieves remarkable performance on resisting different adversarial attacks.
Apart from the improved robustness, we also observe increased inference accuracies on the benign test-sets (cf., the fourth column of Table 1 in the paper~\cite{Yan2018}).
We believe the superiority of our inference accuracies is related to the nature of large margin principle.
In order to analyze the conjecture in detail, we first formalize the definition of an instance-specific margin and introduce some prior theoretical results~\cite{Xu2012} as below.

\begin{definition}\label{def:margin}
 Let us denote by $g: \mathbb R^m \rightarrow \{\pm1\}$ a decision function, then the instance-specific margin $\gamma^p_{\mathbf x}$ of a sample $\mathbf x \in \mathbb R^m$ w.r.t. $g$ is the minimal $\ell_p$ distance from $\mathbf x$ to the decision boundary.
\end{definition}

\begin{definition}\cite{Xu2012}\label{def:robustness}
 Let $\mathcal S_n$ be a sampled training set, and $l$ be the loss function. An algorithm is $(K, \epsilon (\mathcal S_n))$-robust if the sample space $\mathcal S=\mathcal X\times \mathcal Y$ can be partitioned into K disjoint sets denoted by $\mathcal K_k$, $k=1,\ldots,K$, such that for all $s_j=(\mathbf x_j, y_j)\in \mathcal S_n$ and $s= (\mathbf x,y) \in \mathcal S$,
 \begin{equation}
  s_j , s \in \mathcal K_k \ \Rightarrow \ |l(g(\mathbf x_j), y_j)-l(g(\mathbf x), y)| \leq \epsilon (\mathcal S_n).
 \end{equation}
\end{definition}

\begin{theorem}\cite{Xu2012}\label{theo:margin_robustness}
 If there exists $0< \gamma^p <\gamma^p_{\mathbf x_j}$ for all $j$, then the learning algorithm is $(2\mathcal N (\gamma^p/2 , \mathcal X; d_p), 0)$-robust, in which $\mathcal N (\gamma^p/2 , \mathcal X; d_p)$ is the $\epsilon$-covering number\footnote{The definition of $\epsilon$-covering number can be found in~\cite{Van1996}, and it is monotonically decreasing w.r.t. its first argument.} of the input space $\mathcal X$, and $d_p$ is the $\ell_p$ norm, in which the 0/1 loss $l(g(\mathbf x_j), y_j)=\mathbf 1(g(\mathbf x_j)\neq y_j)$ is chosen.
\end{theorem}

Theorem~\ref{theo:margin_robustness} establishes an intrinsic connection between the concerned instance-specific margin and a defined ``robustness''.
Such robustness is different from the adversarial robustness by definition, but it theoretically connects to the generalization error of $g$, as shown in Theorem~\ref{theo:robustness_ge}.

\begin{theorem}\cite{Xu2012}\label{theo:robustness_ge}
 Let $\mu$ be the underlying distribution of the sample $(\mathbf x, y)$. If an algorithm is $(K, \epsilon(\mathcal S_n))$-robust and $l(g(\mathbf x), y)\leq M$ for all $s=(\mathbf x, y)\in \mathcal S$, then for any $\delta>0$, with probability at least $1-\delta$, it holds that
 \begin{equation}
  \mathrm{GE}(g)\leq \epsilon(\mathcal S_n)+M\sqrt{\frac{2K\log(2)+2\log(1/\delta)}{n}},
 \end{equation}
 in which $\mathrm{GE}(g)$ is the generalization error of $g$, given by
 \begin{equation}
  \mathrm{GE}(g)=\left\lvert \mathbb E_{(\mathbf x, y)\sim\mu} l(g(\mathbf x), y)-\frac{1}{n}\sum_{(\mathbf x_j, y_j)\in \mathcal S_n}l(g(\mathbf x_j), y_j)\right\rvert.
 \end{equation}
\end{theorem}

Theorem~\ref{theo:robustness_ge}, along with Theorem~\ref{theo:margin_robustness}, advocates a large margin in the input space and guarantees the generalization ability of learning machines.
Also, it partially explains the superiority of our Deep Defense trained DNNs on benign-set inference accuracies.
However, since the regularizer $L_{\mathrm {DD}}$ is originally designed for resisting attacks, it may be suboptimal for improving the generalization ability (or reducing the generalization error) of learning machines.

In fact, for linear binary classifiers where $f(\mathbf{x})=\mathbf{w}^T\mathbf{x}+b$, assuming the training data is fully separable, then the regularization boils down to minimizing\footnote{We choose $\lambda=c=d=1$, and $R(t)=t$ for simplicity.}:
\begin{equation}\label{eq:binlinreg0}
 \sum_{k\in \mathcal T}-\frac{\left\lvert\mathbf{w}^T\mathbf{x}_k+b\right\rvert}{\lVert\mathbf{w}\rVert_2}+\sum_{k\in\mathcal F}\frac{\left\lvert\mathbf{w}^T\mathbf{x}_k+b\right\rvert}{\lVert\mathbf{w}\rVert_2}.
\end{equation}
Since scaling $(\mathbf w, b)$ by any positive scalar $t$ does not change the value of our regularization term, we constrain $\lVert \mathbf w\rVert_2=1$ to make the problem well-posed.
We denote the index set of samples from positive class and negative class by $\mathcal S_n^+$ and $\mathcal S_n^-$, respectively, and further assume the number of training samples in positive and negative classes to be identical (\ie, $\lvert \mathcal S_n^+\rvert=\lvert \mathcal S_n^-\rvert$), then Eq.~\eqref{eq:binlinreg0} can be rewritten as:
\begin{equation}\label{eq:binlinreg1}
 \mathbf w^T\left(\sum_{k\in\mathcal S_n^-}\mathbf x_k-\sum_{k\in\mathcal S_n^+}\mathbf x_k\right),
\end{equation}
in which the bias term $b$ has been canceled out.
Obviously, minimizing Eq.~\eqref{eq:binlinreg1} under the constraint $\lVert \mathbf w\rVert_2=1$ yields $\mathbf w^*=(\sum_{k\in\mathcal S_n^+}\mathbf x_k-\sum_{k\in\mathcal S_n^-}\mathbf x_k)/Z$, where $Z$ is a normalizer to make sure $\lVert \mathbf w^*\rVert_2=1$.
Geometrically, the decision boundary corresponding to calculating $\mathbf w^*$ is orthogonal to the line segment connecting the centers of mass of positive training samples and that of negative training samples.
Note that all training samples in~(\ref{eq:binlinreg1}), no matter how far away from the decision boundary, have equal contribution to $\mathbf w^*$.
An undesired consequence of such formulation is that, the regularizer can be severely influenced by samples not really close to the decision boundary.
As a result, such $\mathbf w^*$ may process a poor global margin $\gamma^p$, since $\gamma^p$ is generally the ``worst-case'' distance from training samples to the decision boundary, dominated mainly by those close to it.

\begin{figure*}[tp]
 \centering
 \newcommand{\subfigwidth}{0.24\linewidth}
 \subfloat[\textsc{AVG+LIN}]{\includegraphics[width=\subfigwidth]{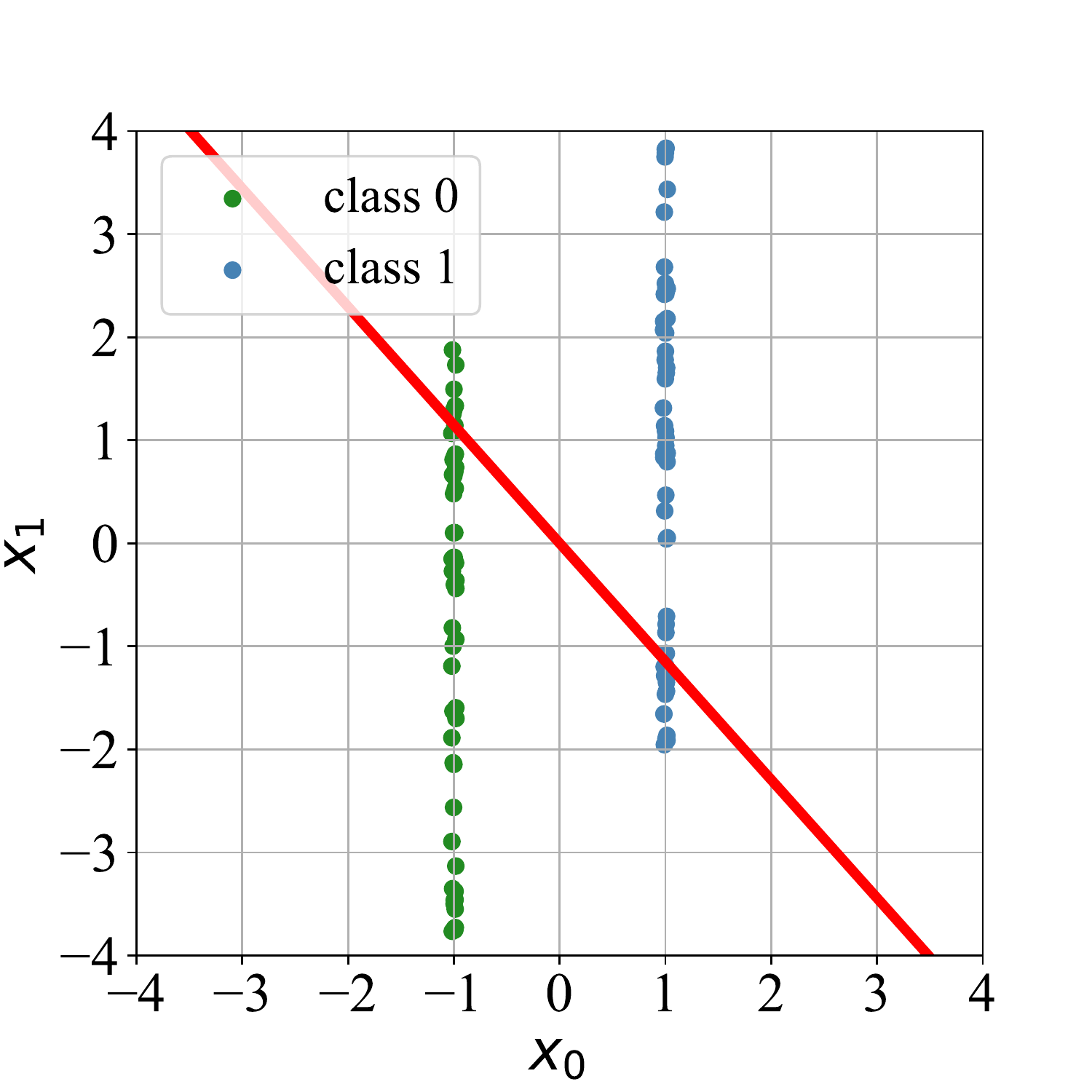}}
 \subfloat[\textsc{AVG+EXP}]{\includegraphics[width=\subfigwidth]{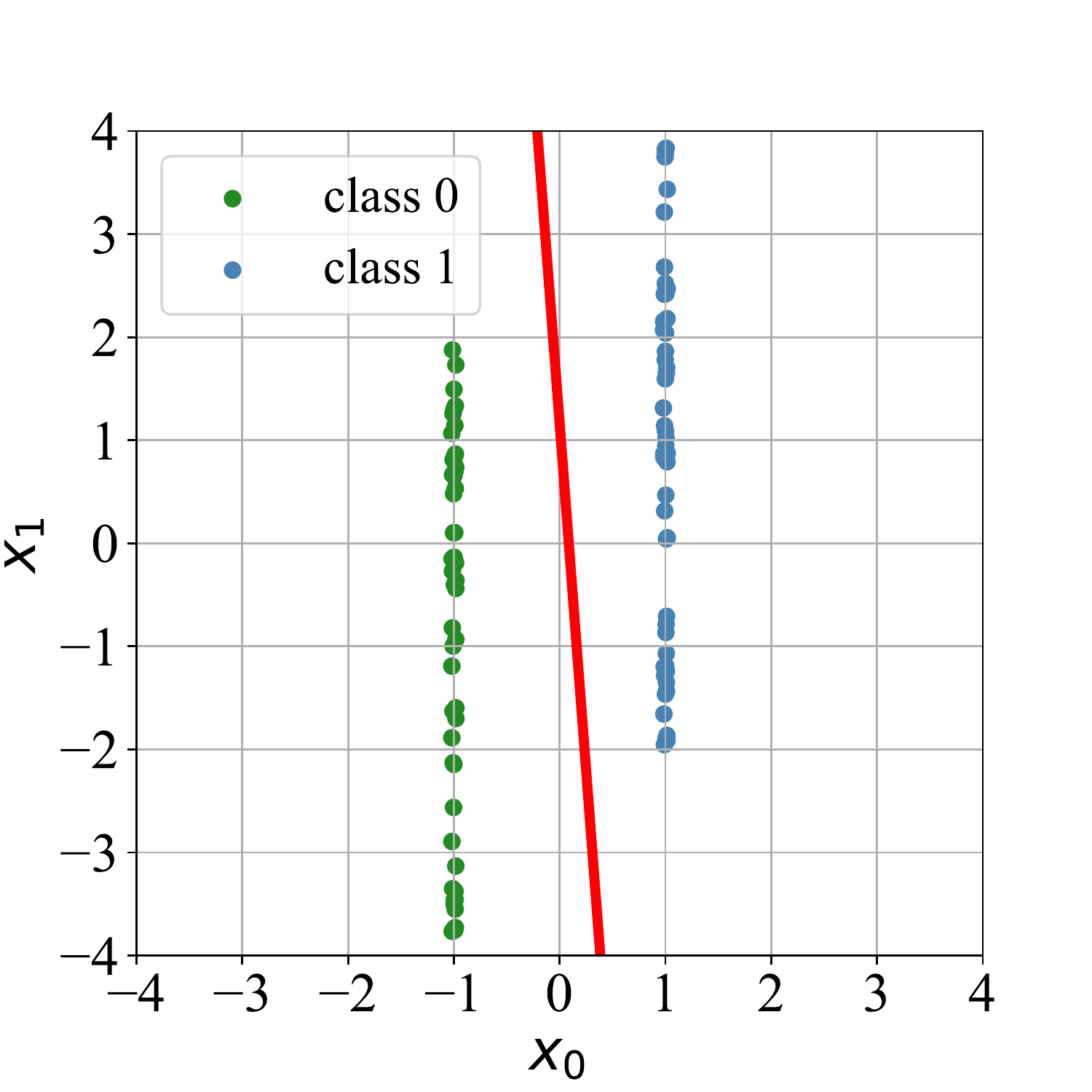}}\
 \subfloat[\textsc{AVG+INV}]{\includegraphics[width=\subfigwidth]{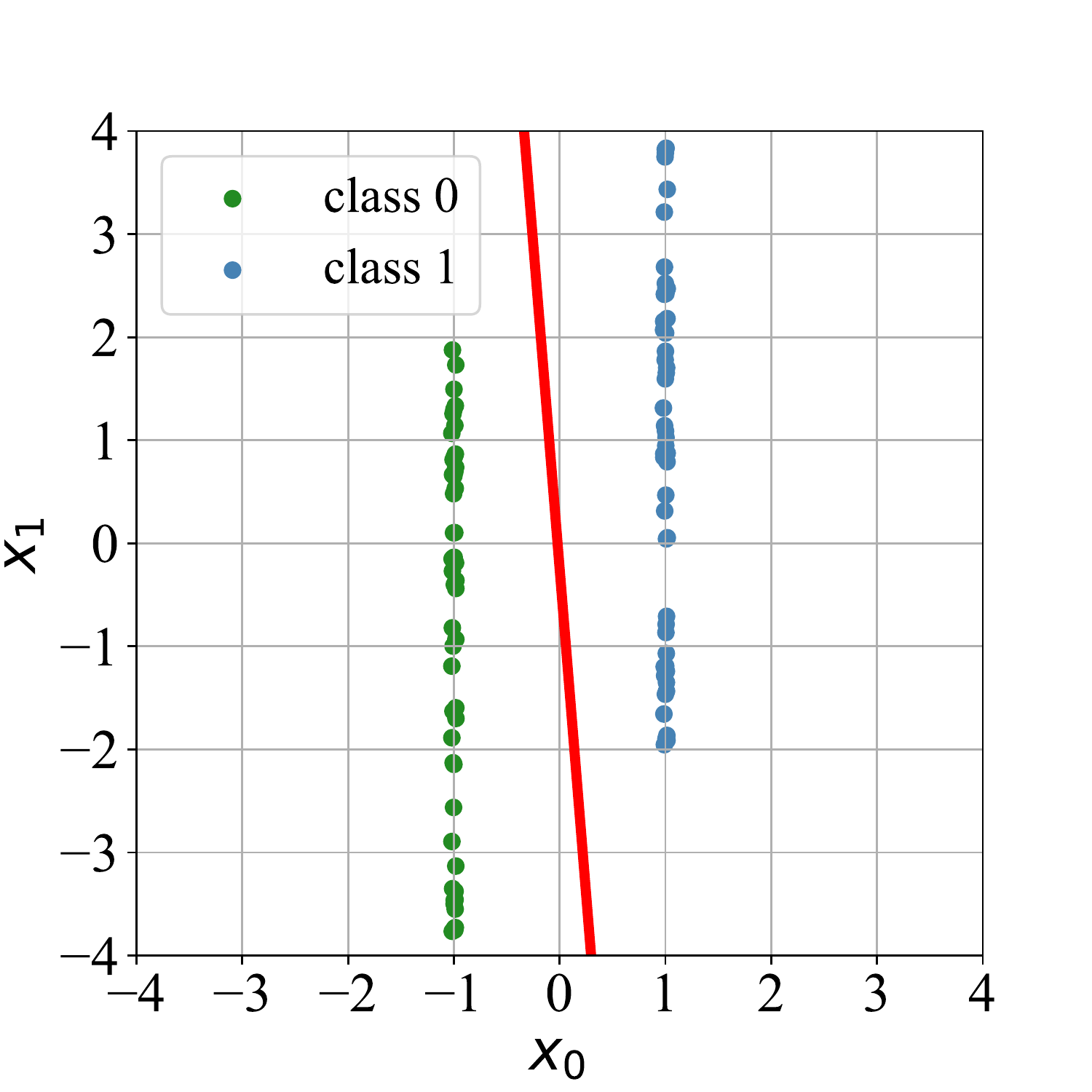}}
 \subfloat[\textsc{MIN+LIN/EXP/INV}]{\includegraphics[width=\subfigwidth]{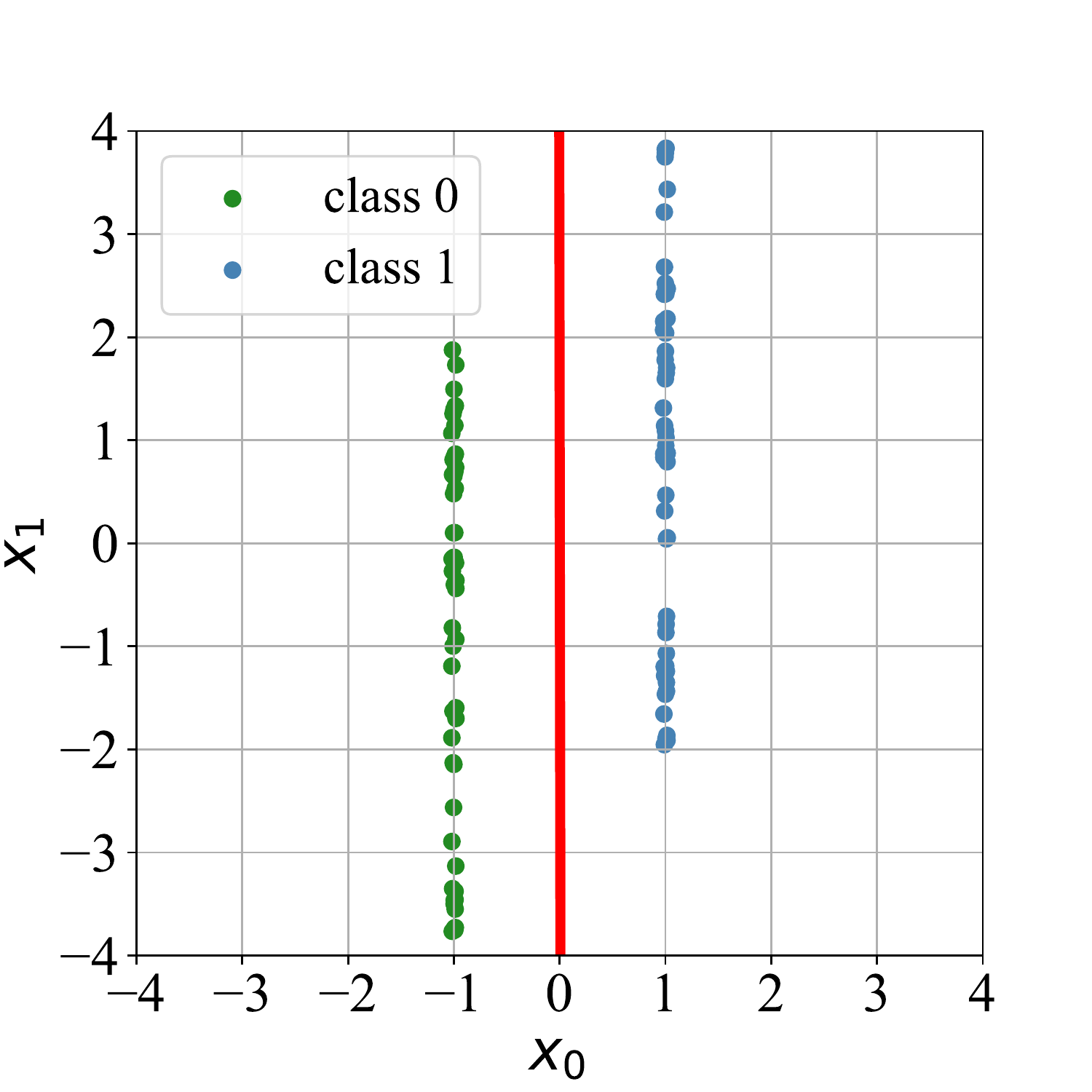}}

 \caption{Different AMM configurations on synthesized 2D data for linear classifiers. (a): The \textsc{AVG+LIN} configuration results in a separating line which is orthogonal to the line connecting the centers of different classes. (b), (c): By adjusting the scales of regularization term of different data points using our non-linear shrinkage functions, we obtain decision boundaries with improved margins and better generalization ability compared with the linear shrinkage function adopted in \textsc{AVG+LIN}. (d): When the regularization is only applied to samples with small (instance-specific) margins, the learned models almost show optimal performance up to the limit on numerical precisions. The decision boundaries are plotted in red. Best viewed in color.}
 \label{fig:reg_configs}
\end{figure*}

\subsection{Our Main Framework}\label{sec:large_margin_dnns}
We know from the previous section that although the margin, robustness and generalization ability of DNNs are theoretically connected, there is an intrinsic distinction between our current method and a desired margin maximization.
In this section, we provide further analyses and introduce aggregation function (in Section~\ref{sec:agg_fun}) and shrinkage function (in Section~\ref{sec:shrink_fun}) designed specifically to exploit the large margin principle more effectively in practice.

\subsubsection{Aggregation Function}\label{sec:agg_fun}
Deep Defense in Eq.~\eqref{eq:regularizer1} aggregates regularization information from training samples (in a batch) by taking average.
However, this aggregation strategy can be suboptimal for improving the generalization ability, while Theorem~\ref{theo:robustness_ge} also suggests a minimal perturbation (rather than the adopted average).
Ideally, one can apply regularization only on the sample with minimal perturbation over the whole training set to maximize $\gamma^p=\min_j\gamma^p_{\mathbf x_j}$.
Nevertheless, the gradient of such a regularizer will be zero for most of the training samples, and in practice it takes much longer time to train and achieve satisfactory results.
\emph{This is different from the well-known scheme in linear SVM.}

To achieve a reasonable trade-off between the theoretical requirement and regularization strength, we consider using a \textsc{min} aggregation function within each batch instead of the whole training set during training.
Specifically, for correctly classified samples $\mathbf x_k$ we apply regularization to it iff two conditions are fulfilled simultaneously: (a). $\lVert\Delta_{\mathbf x_k}\rVert$ is the smallest among all samples with the same ground-truth label in this batch, and (b). $\lVert\Delta_{\mathbf x_k}\rVert$ belongs to the top 20\% smallest in this batch.
If a correctly classified sample does not satisfy these two conditions, we simply set its regularization term to zero in the current training step.
While if a sample is misclassified by the current model,
we expect to decrease its distances to the correct predictions.
Analogous to the above codec for \textsc{MIN}, we denote the original Deep Defense strategy (\ie, averaging all) as \textsc{AVG}.

\subsubsection{Shrinkage Function}\label{sec:shrink_fun}
As discusses~\cite{Yan2018}, if we penalize an perturbation directly (\ie, setting $R(t)=t$ in Eq.~\eqref{eq:regularizer0} and~\eqref{eq:regularizer1}), some ``extremely robust'' samples may dominate the regularization term, which shall pose a negative impact on the training procedure.
What's worse, the regularization term will never diminish to zero with a linear $R(\cdot)$.
To alleviate the problems, we attempt to choose a nonlinear "shrinkage" function for $R(\cdot)$.
It should be monotonically increasing, such that the correctly classified samples with abnormally small $\lVert\Delta_{\mathbf x_k}\rVert$ are penalized more than those with relatively large values.
Essentially, concentrating more on samples with small instance-specific margins coheres with the evidence in Theorem~\ref{theo:robustness_ge}, since we know the minimal (instance-specific) margin probably connects the most to the generalization ability.
We will demonstrate the performance of different choices: (a). $R(t)=t$, denoted by \textsc{LIN}, (b). $R(t)=\exp(t)$, denoted by \textsc{EXP} and (c), $R(t)=\frac{1}{1-t}$, denoted by \textsc{INV}, on training DNNs, \emph{which also differs from the well-known scheme in linear SVM.}
For \textsc{INV}, we make sure $t<1$ by setting appropriate values for $c,d$ and truncating abnormally large values for $\lVert\Delta_{\mathbf x_k}\rVert$ with a threshold.

\subsubsection{Experiments on Toy Data}\label{sec:exp_toy_data}
We first conduct an explanatory experiment by synthesizing 2D data to illustrate how the choices of the functions may affect classification in a binary case.
Suppose that the 2D data from the two classes are uniformly distributed in rectangles $[-1.01, -4.00; -0.99, 2.00]$ and $[0.99, -2.00; 1.01, 4.00]$, respectively.
For each class, we synthesize 200 samples for training and another 200 held-out for testing.
We train linear classifiers with $f(\mathbf{x})=\mathbf{w}^T\mathbf{x}+b$ to minimize the regularization term taking various forms.
We set batch size to 20, and train models for 1000 epochs with the standard SGD optimizer.
The learning rate is initially set to 0.1, and cut by half every 250 epochs.
We use the popular momentum of 0.9 and a weight decay of 1e-4.

The learned decision boundaries and in different configurations are illustrated in Fig.~\ref{fig:reg_configs} together with test samples.
With purely \textsc{AVG+LIN}, the obtained boundary is roughly orthogonal to the line connecting (-1.0, -1.0) and (1.0, 1.0), which is consistent with our previous analysis in Section~\ref{sec:lims}.
Although we have striven to penalize incorrect classifications, the obtained model in this setting is still unable to gain excellent accuracy, because all training samples contribute equally to pushing the decision boundary.
Better generalization ability and thus improved test-set accuracies can be obtained in the \textsc{AVG+EXP} and \textsc{AVG+INV} settings, as depicted in Fig.~\ref{fig:reg_configs} (b) and (c).
This verifies the effectiveness of the non-linear shrinkage functions aimed at regularizing "extremely robust" samples less.
It can also be seen that the \textsc{MIN} aggregation function leads to a more sensible margin.
Optimal (or near optimal) boundaries are obtained in the \text{MIN} settings, which attains the largest possible margin of 1.
As discussed in Section~\ref{sec:lims}, the \textsc{MIN} aggregation function is more related with the geometric margin in comparison with the \textsc{AVG} so it can directly facilitate the margin as well as the generalization ability.

\begin{table}[!t]
 \caption{Results on MNIST: Compare MLP Models Trained Using Different Aggregation and Shrinkage Functions}\label{tab:mlp_agg_shrink}
 \centering
 \begin{tabular}{lccc}
  \toprule
  Method   & Aggregation  & Shrinkage    & Error Rate (\%)    \\
  \midrule
  baseline & -            & -            & 1.79$\pm$0.06      \\ \hline
           & \textsc{AVG} & \textsc{LIN} & 2.26$\pm$0.05      \\
           & \textsc{AVG} & \textsc{INV} & 1.18$\pm$0.03      \\
  AMM      & \textsc{AVG} & \textsc{EXP} & 0.94$\pm$0.02      \\
           & \textsc{MIN} & \textsc{LIN} & 1.28$\pm$0.02      \\
           & \textsc{MIN} & \textsc{INV} & 0.97$\pm$0.03      \\
           & \textsc{MIN} & \textsc{EXP} & \bf{0.90$\pm$0.03} \\ \bottomrule
 \end{tabular}
\end{table}

\begin{figure}[t]
 \centering
 \vskip -0.2in
 \includegraphics[width=0.95\linewidth]{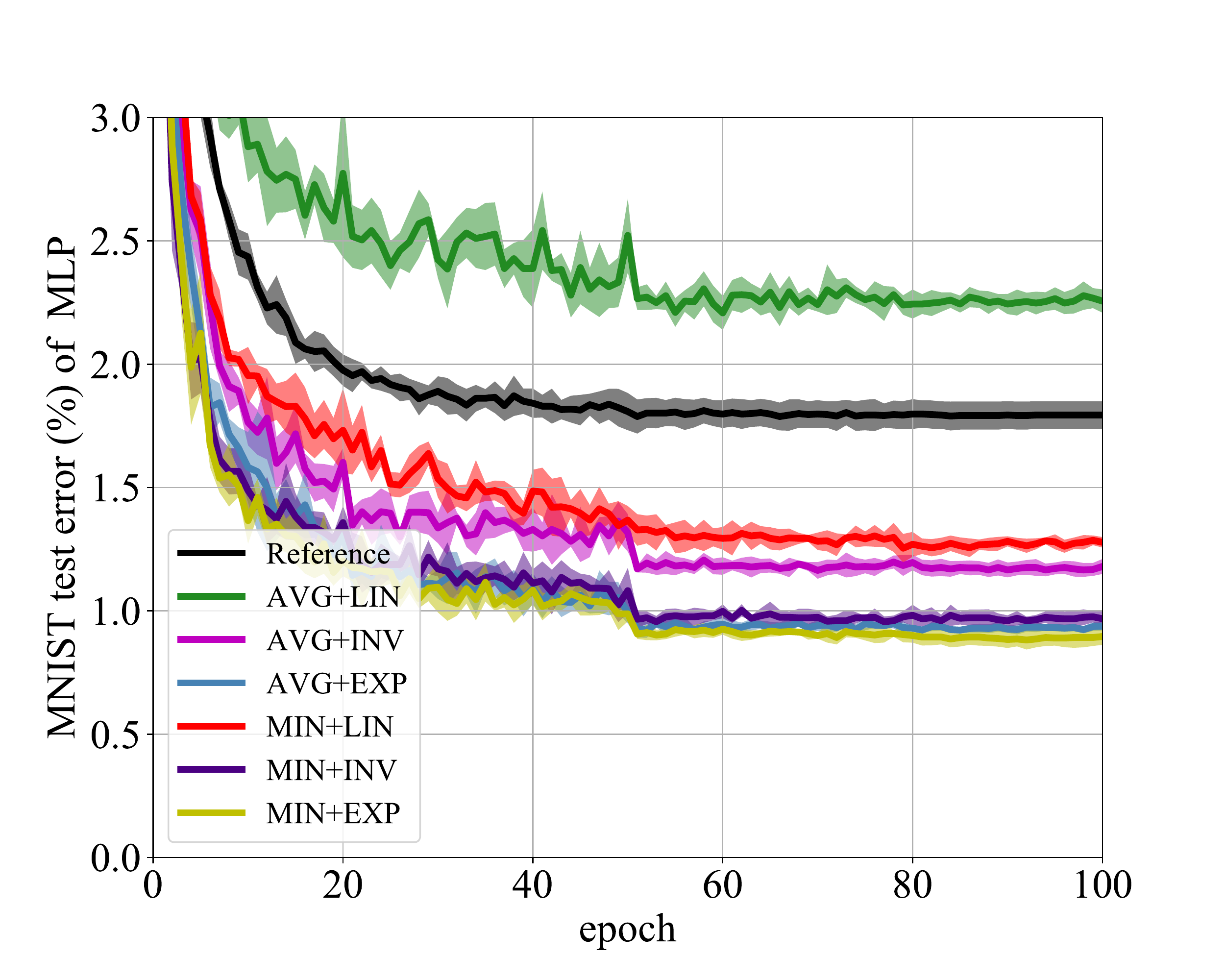}
 \vskip -0.1in
 \caption{Convergence curves of MLP models trained using our AMM with different aggregation and shrinkage functions on MNIST.}
 \label{fig:mlp_agg_shrink}
 \vskip -0.1in
\end{figure}

We further conduct an experiment on MNIST~\cite{Lecun1998}, which is a real-world dataset.
Here a simple multilayer perceptron (MLP) is adopted and trained with cross-entropy loss.
To achieve the best performance within each configuration, we first run a grid search for $\lambda$, $c$, and $d$.
Table~\ref{tab:mlp_agg_shrink} shows the final error rates while Fig.~\ref{fig:mlp_agg_shrink} illustrates the convergence curves of our AMM with different aggregation and shrinkage functions.
We repeat the training five times with different initialization instantiations to report also the standard derivations of error rates.
We see with \textsc{AVG+LIN} the obtained mean error rate even increases from 1.79\% to 2.29\% in comparison with the reference model, proving that treating all samples equally in the regularizer can pose negative impact on the generalization ability.
With the help of the \textsc{MIN} aggregation function, our AMM achieves a 1.28\% error rate, which is far lower than that with \textsc{AVG+LIN} (2.26\%), as well as that of the reference model (1.79\%).
Such positive effect of the \textsc{MIN} aggregation function is consistent with our observation on the synthetic 2D data.

The benefit of the nonlinear shrinkage functions is also highlighted.
When compared with the \textsc{LIN} function, nonlinear shrinkage functions \textsc{INV} and \textsc{EXP} gain relative error decreases of 47\% and 58\% within the same \textsc{AVG} setting, and 24\% and 29\% within the \textsc{MIN} setting.
Such results well explain our intuitions and insights in Section~\ref{sec:shrink_fun}.

\subsubsection{Implementation Details and Discussions}\label{sec:method_discussion}
Our framework has an intrinsic connection with SVM.
For a linear binary classification problem with separable training data, SVM can be viewed as a special case in our framework provided the model is linear, if we remove the classification objective $L$ in the whole optimization problem~\eqref{eq:whole_opt_problem}.
It can be easily verified that the \textsc{MIN+LIN} regularization (along with the weight decay term) is the equivalent with a hard margin linear SVM, if the current model achieves excellent accuracy (\ie 100\%) on the training set and we use all training samples in each batch.
Moreover, as testified in the previous section, the aggregation function ought to be more essential than the shrinkage function in linear binary cases.

The landscape and margin of nonlinear DNNs are much more complex and generally infeasible to compute in comparison with those of linear models, especially when multiple classes get involved.
Our framework exploits an adversarial perturbation as a proxy of the margin.
With different configurations on the aggregation and shrinkage functions, it formulates a variety of regularization types.
They might devote more to the generalization ability (\eg, the one with \textsc{MIN+EXP}) or robustness (\eg, our Deep Defense with \textsc{AVG} and approximately \textsc{EXP}).
All the encompassed variants share a similar core idea that is to incorporate an adversarial attack into the training process.
In particular, we utilize the perturbation norm as an estimation of the margin.
Current state-of-the-art attacks typically achieve $\sim$100\% success rate on powerful DNNs, while the norm of perturbation can be reasonably small and thus fairly close to the real margin values.
We specifically choose $p=2$ to comply with previous theoretical analysis.
Also, we know from Section~\ref{sec:exp_toy_data} that the \textsc{MIN+EXP} trades off test-set performance in favor of theoretical margins.
We leave the choices for the classification objective $L$ to customized network configurations, in parallel with our AMM configurations.
In fact, we have tested our AMM with popular choices for $L$ including the cross-entropy loss and hinge loss but never found a significant difference in the experiments.

By delving deeply into the geometric margin, we unify a set of learning-based regularizers within the proposed AMM framework.
Guidelines are correspondingly provided in case one prefers the generalization ability to robustness or DNNs to linear models.
Contemporaneous with our work, Elsayed \etal~\cite{Elsayed2018} propose to linearize the forward mapping of DNNs, somewhat similar to a single-step Deep Defense without utilizing the high order gradients (as in Section~\ref{sec:dd}) and nonlinear shrinkage function (as in Section~\ref{sec:shrink_fun}).
See more discussions in Section~\ref{sec:exp_ho}.

\section{Experimental Verifications}\label{sec:exp}
In this section, we experimentally verify the remarks and conjectures presented in previous sections and evaluate the performance of our AMM with specifically \textsc{MIN+EXP} on various datasets (including MNIST, CIFAR-10/100, SVHN and ImageNet).
We compare our derived method with the state-of-the-arts to demonstrate its effectiveness.

\subsection{Datasets and Models}\label{sec:datasets_and_models}
We perform extensive experiments on five commonly used classification dataset: MNIST~\cite{Lecun1998}, CIFAR-10/100~\cite{krizhevsky2009}, SVHN~\cite{Netzer2011} and ImageNet~\cite{Deng2009}.
Dataset and network configurations are described as below.
For MNIST, CIFAR-10/100, and SVHN, we construct a held-out validation set for hyper-parameter selection by randomly choosing 5k images from the training set.
For ImageNet, as a common practice, we train models on the 1.2 million training images, and report top-1 error rates on the 50k validation images.

\subsubsection{MNIST}
MNIST consists of 70k grayscale images, in which 60k of them are used for training and the remaining are used for test.
We train deep networks with different architectures on MNIST: (a). a four-layer MLP (2 hidden layers, 800 neurons in each) with ReLU activations, (b). LeNet~\cite{Lecun1998} and (c). a deeper CNN with 12 weight layers named ``LiuNet''~\cite{Liu2016,Wang2018}.
Similar to many previous works, we subtract the mean for both training and test data in pre-processing, and no data augmentation is adopted. For more details about these architectures, please see our appendix.

\subsubsection{CIFAR-10/100, and SVHN}
Both CIFAR-10 and CIFAR-100 contain 60k color images, including 50k training images and 10k test images.
SVHN is composed of $\sim$630k color images in which $\sim$604k of them are used for training and the remaining for testing.
For these datasets, we train six networks: (a). a light ConvNet with the same architecture as in~\cite{Xie2016}, (b). the network-in-network (NIN)~\cite{Lin2014}, (c). the ``LiuNet`` as applied in the CIFAR-10 experiments in~\cite{Liu2016}., (d)(e). the standard ResNet-20/56~\cite{He2016} architectures, and (f) a DenseNet-40~\cite{Huang2017} in which all layers are connected.
We uniformly resize each image to 36x36, and randomly crop a 32x32 patch during training as data augmentation.
Moreover, we apply random horizontal flipping with a probability of 0.5 to combat overfitting, except SVHN.

\subsubsection{ImageNet}
ImageNet is a highly challenging image classification benchmark which consists of millions of high-resolution images over 1,000 classes.
Starting from ResNet~\cite{He2016}, deep models with skip connections have advanced the state-of-the-arts on this highly challenging dataset~\cite{Zhang2017,Huang2017,Xie2017resnext,Hu2018}.
We adopt ResNet-18/50~\cite{He2016} and SENet-50~\cite{Hu2018} which includes numerous skip connections as representative architectures to validate our method.
Following previous works~\cite{Szegedy2015}, we randomly crop a patch whose size is uniformly distributed between 8\% and 100\% of the original image size, with aspect ratio uniformly distributed in $\left[\frac 3 4, \frac 4 3\right]$.
Then we resize the cropped patch to 224x224 and feed it into the network.
As a common practice, random horizontal flipping is also applied.

\subsection{Training Protocol}
We use the cross-entropy loss in the training objective for $L$, as with previous works.
Table 2 in the appendix summarizes the batch size, maximal number of epoch, and learning rate policy used in our experiments.
We start training with some initial learning rate (shown in the 5-th column of Table 2) and we anneal them by some multipliers at certain epochs (specified in the 6-th column).
The standard stochastic gradient descent optimizer with a momentum of 0.9 and a weight decay of 1e-4 is adopted in all experiments.
All the hyper-parameters are tuned on the validation set with reference networks in order to achieve their supreme performance.

For relatively small models and datasets, we initialize models with the so-called ``MSRA'' strategy~\cite{He2015} and train from scratch, otherwise we fine-tune from our trained reference models (more details can be found in Table 2 in the appendix).
To avoid abnormally large gradients and probably a drift away of classification loss, we project gradient tensors onto a Euclidean sphere with radius 10, if their norm exceeds the threshold 10.
This technique is also known as ``clip gradients'' and has been widely adopted in the community.
When calculating adversarial perturbations for our AMM, we allow a maximal iteration of $u=6$, which is sufficient to fool DNNs on $\sim$100\% of the training samples in most cases.
Hyper-parameters $\lambda,c,d$ in our regularizer is determined by cross validation in the held-out validation set, as described.

\subsection{Exploratory Experiments on MNIST}\label{sec:exploratory_exp}
As a popular dataset for evaluating the generalization performance of classifiers~\cite{Tang2013,An2015,Blundell2015,Liu2016,Xie2016}, MNIST is a reasonable choice for us to get started.
We shall analyze the impacts of different configurations in our framework.

\subsubsection{Effect of High Order Gradients and Others}\label{sec:exp_ho}
Let us first investigate the effect of introducing high order gradients, which serves as an essential component in our framework.
It is triggered when back-propagating gradients through $\nabla f$ in our regularizer, which is usually difficult to formalize and compute for DNNs.
We invoke the automatic differentiation mechanism in PyTorch~\cite{Pytorch2017} to achieve this.
One can also build inverse networks to mimic the backward process of DNNs, as in~\cite{Yan2018}.

\begin{figure}[t!]
 \centering
 \includegraphics[width=0.95\linewidth]{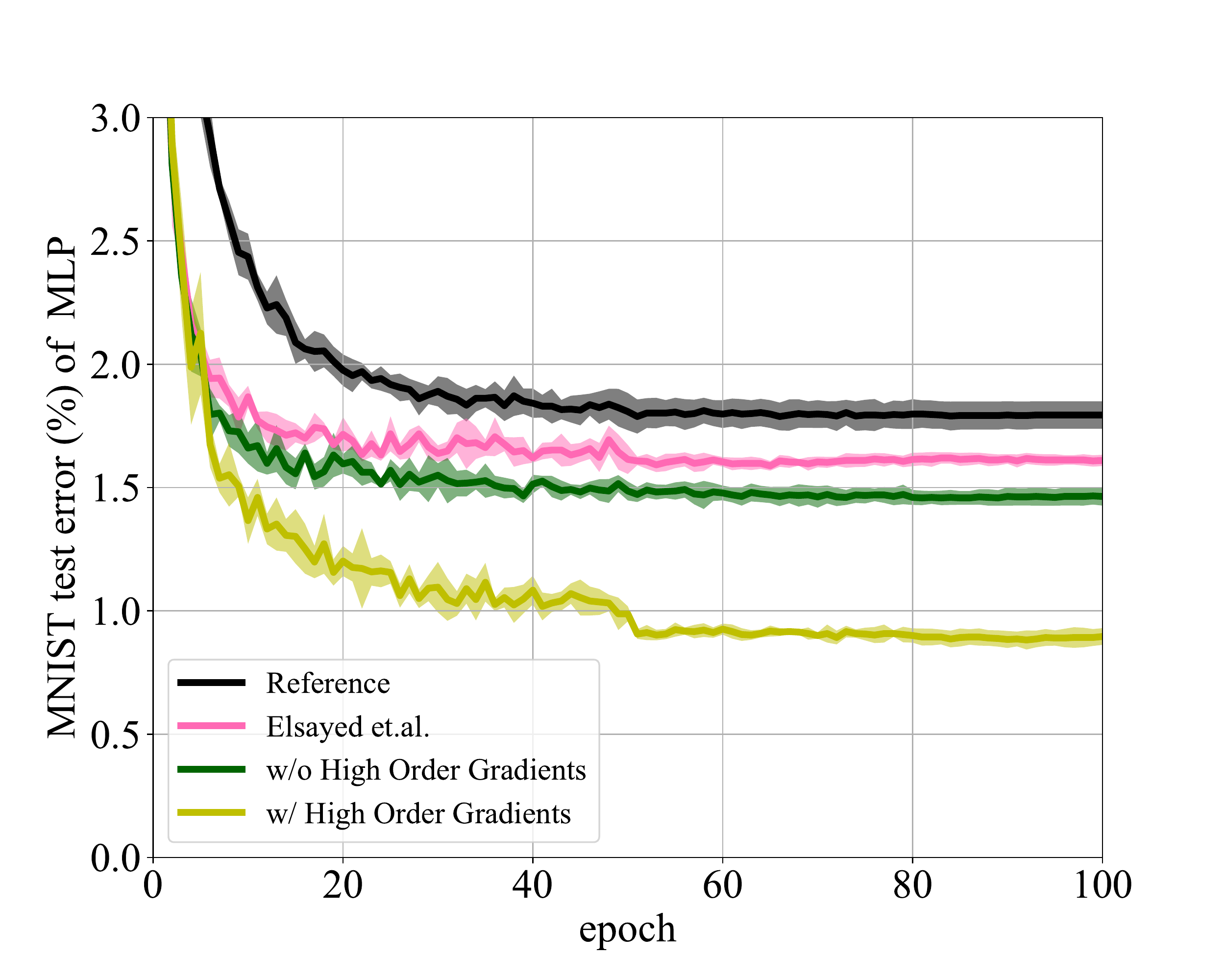}
 \caption{Convergence curves of MLP models on MNIST with or w/o high order gradients.}
 \label{fig:mlp_hograd}
\end{figure}

\begin{table}[!t]
 \caption{MNIST Test Error Rates of MLP Models w/ or w/o High Order Gradients}\label{tab:mlp_hograd}
 \centering
 \begin{tabular}{cc}
  \toprule
  Model                    & Error (\%)         \\
  \midrule
  Reference                & 1.79$\pm$0.06      \\
  w/o high order gradients & 1.46$\pm$0.04      \\
  w/ high order gradients  & \bf{0.90$\pm$0.03} \\ \bottomrule
 \end{tabular}
\end{table}

We try masking the gradient flow of $\nabla f$ by treating the entries of $\Delta_{\mathbf{x}}$ as constants, as done by Elsayed \etal~\cite{Elsayed2018}.
In general, they expect to enlarge the margin by penalizing the $\ell_2$ norm of a linear perturbation, if it goes below a threshold.
Such approximation may lead to conceptually easier implementations but definitely also results in distinctions from the gradient direction to pursue a large margin.
With or without the high order gradients, we obtain different MLP models using our AMM.
They are compared in Table~\ref{tab:mlp_hograd} and Fig.~\ref{fig:mlp_hograd}, along with the ``Reference'' that indicates the baseline MLPs with $\lambda=0$ (\ie, no AMM).
Means and standard deviations of the error rates calculated from all five runs are shown.

Though both methods achieve decreased error rates than the ``Reference'', models trained with gradient masked on $\nabla f$ (\ie, without high order gradients) demonstrate apparently worse performance (1.46\%, pink in Fig.~\ref{fig:mlp_hograd}) than those with full gradients (0.90\%, yellow in Fig.~\ref{fig:mlp_hograd}).
Except for the high order gradients, Elsayed \etal's method~\cite{Elsayed2018} also miss several other components that may further hinder it from achieving comparable performance with ours.
For empirical validations, we try following its main technical insights and implementing the method for empirical validation.
We follow the single-step setting and a threshold-based shrinkage function and summarize its results in Table~\ref{tab:mlp_hograd} and Fig.~\ref{fig:mlp_hograd} for comparison.
We see its prediction error is even higher than ours without high order gradients.

\subsubsection{Training with Less Samples}
Our AMM enhances the generalization ability and robustness of DNNs in different aspects.
We consider the possibility of training DNN models with fewer samples in this section.
Specifically, we sample $\{5\mathrm{k},10\mathrm{k},15\mathrm{k},\ldots,60\mathrm{k}\}$ images randomly from the MNIST training set, and train MLP models on these subsets.
Once sampled, these subsets are fixed for all training procedures.
The \textsc{MIN+EXP} setting and identical $\lambda$, $c$, and $d$ as in our previous experiments are adopted.

\begin{figure}[t!]
 \centering
 \includegraphics[width=0.95\linewidth]{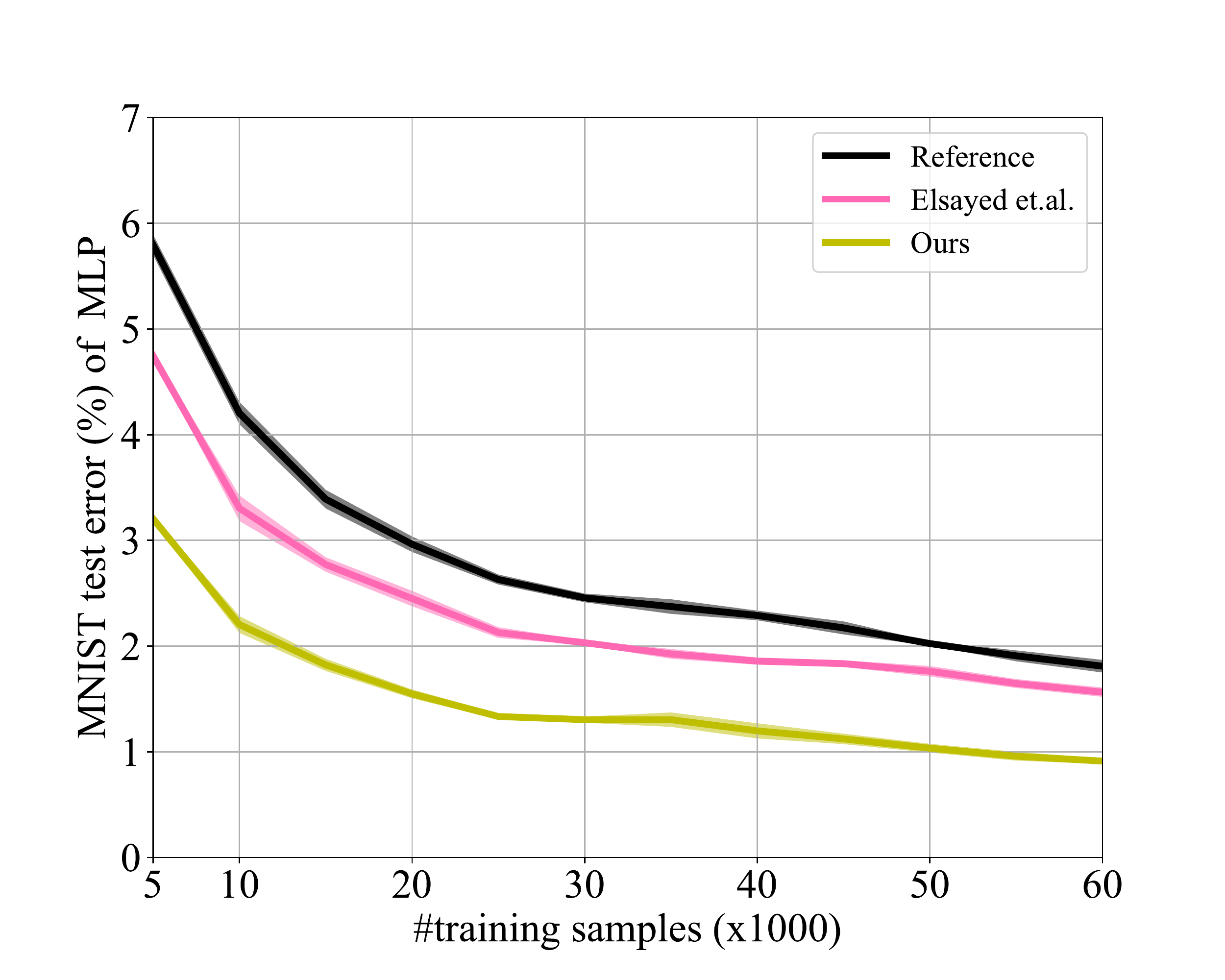}
 \caption{MNIST test error rates of MLP models with less training images.}
 \label{fig:mlp_less}
\end{figure}

Fig.~\ref{fig:mlp_less} illustrates how test error rates vary with the number of training samples.
Same with previous experiments, we perform five runs for each method and the shaded areas demonstrate the standard deviations.
Clearly, models with our full gradient AMM achieve consistently better performance than the ``Reference'' models and other competitors.
With only 20k training images, our method helps to achieve $1.54\pm0.04\%$ error, which is even slightly lower than that of the vanilla models with all 60k training images ($1.81\pm0.05\%$ error).

\subsubsection{Training with Noisy Labels}
We train models with possibly noisy labels in this experiment to simulate unreliable human annotations in many real-world applications.
We shall use all 60k MNIST training images, but for a portion of them, we substitute random integers in [0, 9] for their ``ground-truth'' labels.
For $n\in\{0\mathrm{k}, 5\mathrm{k},10\mathrm{k},\ldots,55\mathrm{k}\}$, we construct 12 training partitions each consists of $n$ images with the original labels and $60\mathrm{k}-n$ images with random labels.

\begin{figure}[t!]
 \centering
 \includegraphics[width=0.95\linewidth]{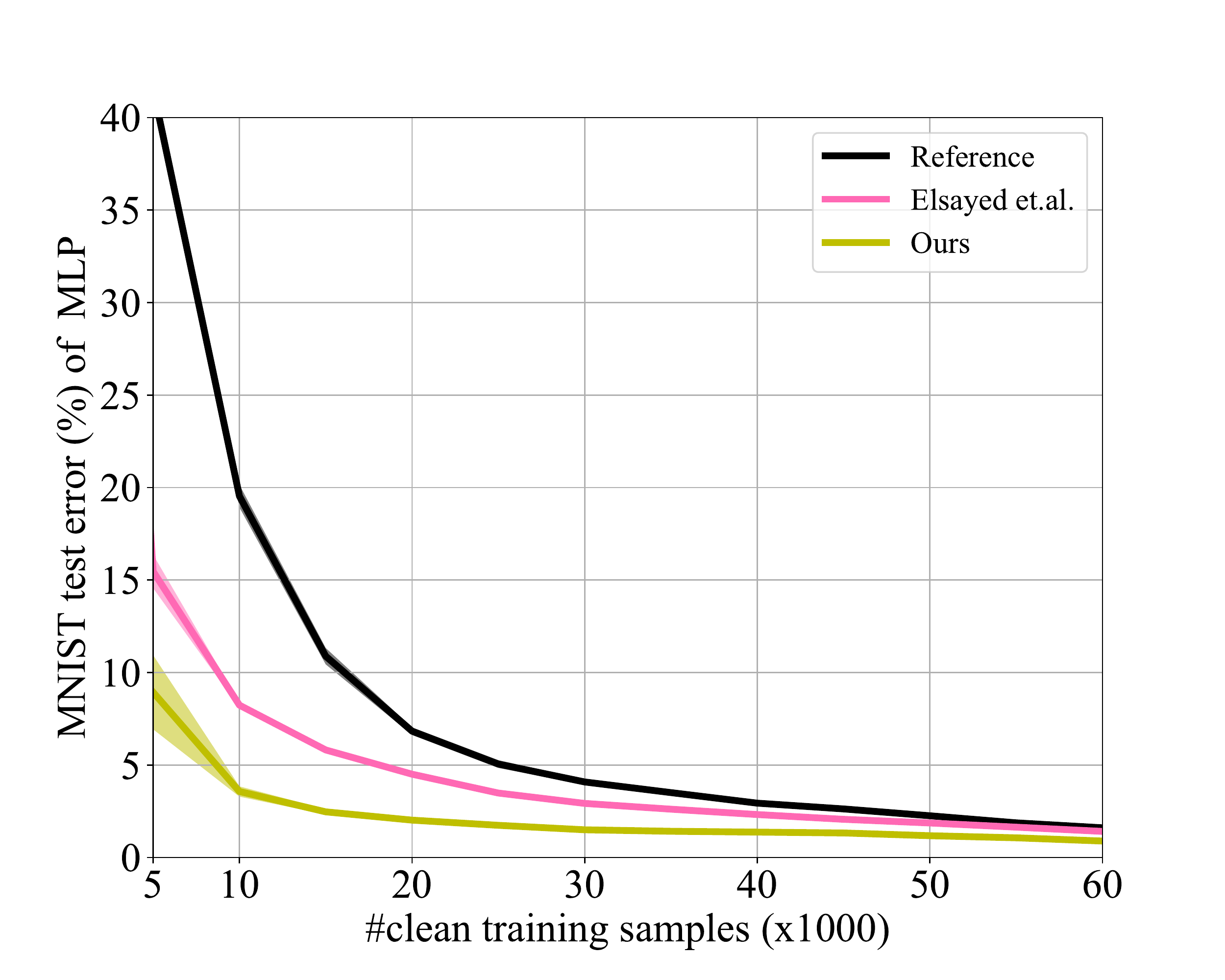}
 \caption{MNIST test error rates of MLP models with noisy training images.}
 \label{fig:mlp_noisy}
\end{figure}

The ``Reference'' and our AMM models are trained, and their error rates are shown in Fig.~\ref{fig:mlp_noisy}.
Without specific regularization, error rates of reference models increase drastically when a portion of labels are corrupted.
For instance, when training on a set containing 55k images with random labels, models with our adversarial regularization are still able to achieve an average test error less than 10\%.
However, that of the reference models goes above 40\%, which are obviously too high, considering it is a 10-class classification problem.
Our implementation of Elsayed\etal's method also achieves promising performance on the test set, though consistently inferior to ours.

\subsection{Image Classification Experiments}
In this section, we testify the effectiveness of our method on benchmark DNN architectures on different image classification datasets including MNIST, CIFAR-10/100, SVHN and ImageNet.
It is compared with a variety of state-of-the-art margin-inspired methods.
Same with previous experiments in Section~\ref{sec:exploratory_exp}, during training we adopt the \textsc{MIN+EXP} regularizer in our method for all experiments.
For evaluation, both error rates and margin (estimated using DeepFool perturbation) on the test set are reported to verify the effectiveness of our method.

\begin{table}[!t]
 \centering
 \resizebox{1.0\linewidth}{!}{
  \begin{threeparttable}
   \caption{MNIST Test Error Rates}\label{tab:mnist}
   \begin{tabular}{ccccc}
    \toprule
    Method                                & Architecture & Error (\%)     & Margin & Augmentation \\
    \midrule
    Bayes by Backprop~\cite{Blundell2015} & MLP (800)    & 1.32           & -      & -            \\
    DropConnect~\cite{Wan2013}            & MLP (800)    & 1.20$\pm$0.03  & -      & -            \\
    DLSVM~\cite{Tang2013}                 & MLP (512)    & 0.87           & -      & Gaussian     \\
    CRN~\cite{An2015}                     & LeNet        & 0.73           & -      & -            \\
    DropConnect~\cite{Wan2013}            & LeNet        & 0.63$\pm$0.03  & -      & -            \\
    DisturbLabel~\cite{Xie2016}           & LeNet        & 0.63           & -      & -            \\
    L-Softmax~\cite{Liu2016}              & LiuNet       & 0.31           & -      & -            \\
    EM-Softmax~\cite{Wang2018}            & LiuNet       & 0.27~\tnote{a} & -      & -            \\
    \midrule
    Reference                             & MLP (800)    & 1.79$\pm$0.06  & 0.76   & -            \\
    Ours                                  & MLP (800)    & 0.90$\pm$0.03  & 1.90   & -            \\
    Reference                             & LeNet        & 0.87$\pm$0.02  & 1.07   & -            \\
    Ours                                  & LeNet        & 0.56$\pm$0.02  & 2.21   & -            \\
    Reference                             & LiuNet       & 0.41$\pm$0.02  & 1.60   & -            \\
    Ours                                  & LiuNet       & 0.33$\pm$0.03  & 3.41   & -            \\ \bottomrule
   \end{tabular}
   \begin{tablenotes}
    \item[a] Ensemble of 2 LiuNet models are used.
   \end{tablenotes}
  \end{threeparttable}}
\end{table}

\subsubsection{MNIST}~\label{sec:exp_mnist}
Aforementioned MLP, LeNet and LiuNet architectures are used as reference models.
Generally, DNNs have to be deterministic during the attack process, such that we can find a reasonable approximation of minimal distance to the fixed decision boundaries.
However, for DNNs equipped with batch normalization layers, if we implement the DeepFool attack na\"ively, the perturbation of a particular sample may depend on other samples in the same batch since all of them share the same mean and variance in batch normalization procedure.
In order to bypass such dependency and achieve better efficiency, our implementation follows that described in~\cite{Liu2016}, with one exception that we replace all batch normalization layers with group normalization~\cite{Wu2018} with group size 32.
Empirically we found this difference has little (often negative) impact on the error rates of reference network.
We adopt the \textsc{MIN+EXP} setting, and use $\lambda$, $c$, $d$ selected on MLPs for all the three architectures for simplicity, although potential better hyper-parameters may be obtained by running grid search on each of them.
The error rates of different methods are shown in Table~\ref{tab:mnist}.
For a fair comparison, we also provide architectures in the second column of Table~\ref{tab:mnist}.
The annotation MLP ($n$) represents an MLP model with two hidden layers, and each of them has $n$ neurons.
Our method outperforms competitive methods considering the MLP and LeNet architecture, except one case where comparisons are not completely fair: DLSVM~\cite{Tang2013} obtains 0.87\% error (our: 0.90\%) using MLP with additional Gaussian noise added to the input images during training, but we do not use any data augmentation techniques,.
For LiuNet, our method also achieves comparable error rate (0.33\%) with L-Softmax~\cite{Liu2016} (0.31\%).
EM-Softmax~\cite{Wang2018} achieves the lowest 0.27\% error using an ensemble of 2 LiuNets, while our performance is measured on a single model.
Moreover, we see our method significantly and consistently decreases the error rates of reference models, by relative improvements of 49\%, 35\%, and 19\% on MLP, LeNet, and LiuNet, respectively.

\begin{table}[!t]
 \centering
 \resizebox{1.0\linewidth}{!}{
  \begin{threeparttable}
   \caption{CIFAR-10 Test Error Rates}\label{tab:cifar10}
   \begin{tabular}{ccccc}
    \toprule
    Method                      & Architecture & Error (\%)     & Margin & Augmentation    \\
    \midrule
    DropConnect~\cite{Wan2013}  & LeNet        & 18.7           & -      & -               \\
    DisturbLabel~\cite{Xie2016} & LeNet        & 14.48          & -      & hflip \& crop   \\
    DLSVM~\cite{Tang2013}       & LeNet        & 11.9           & -      & hflip \& jitter \\
    CRN~\cite{An2015}           & VGG-16       & 8.8            & -      & hflip \& crop   \\
    L-Softmax~\cite{Liu2016}    & LiuNet       & 5.92           & -      & hflip \& crop   \\
    EM-Softmax~\cite{Wang2018}  & LiuNet       & 4.98~\tnote{a} & -      & hflip \& crop   \\
    \midrule
    Reference                   & LeNet        & 14.93          & 0.16   & hflip \& crop   \\
    Ours                        & LeNet        & 13.87          & 0.24   & hflip \& crop   \\
    Reference                   & NIN          & 10.39          & 0.21   & hflip \& crop   \\
    Ours                        & NIN          & 9.87           & 0.30   & hflip \& crop   \\
    Reference                   & LiuNet       & 6.25           & 0.15   & hflip \& crop   \\
    Ours                        & LiuNet       & 5.85           & 0.29   & hflip \& crop   \\
    Reference                   & ResNet-20    & 8.20           & 0.10   & hflip \& crop   \\
    Ours                        & ResNet-20    & 7.62           & 0.22   & hflip \& crop   \\
    Reference                   & ResNet-56    & 5.96           & 0.16   & hflip \& crop   \\
    Ours                        & ResNet-56    & 5.75           & 0.32   & hflip \& crop   \\
    Reference                   & DenseNet-40  & 5.75           & 0.11   & hflip \& crop   \\
    Ours                        & DenseNet-40  & 5.61           & 0.18   & hflip \& crop   \\
    \bottomrule
   \end{tabular}
   \begin{tablenotes}
    \item[a] Ensemble of 2 LiuNet models are used.
   \end{tablenotes}
  \end{threeparttable}}
\end{table}

\subsubsection{CIFAR-10}
For CIFAR-10, we evaluate our method with LeNet, NIN, LiuNet, ResNet-20/56, and DenseNet-40. 
The architecture of LeNet and NIN are directly copied from our previous work~\cite{Yan2018}, and that of LiuNet is adapted from~\cite{Liu2016}.
We choose the CIFAR-10 LiuNet architecture as described in~\cite{Liu2016}, and replace all batch normalization layers with group normalization layers of group size 32, as in our MNIST experiments.
For ResNets and DenseNets, we adopt the standard architectures as described in previous works~\cite{He2016,Huang2017}, and simply freeze all batch normalization layers during both training and testing to break the the inter-batch dependency as described in Section~\ref{sec:exp_mnist}.
Hyper-parameters $\lambda$, $c$, and $d$ are casually tuned on the hold-out validation set as described in Section~\ref{sec:datasets_and_models}, and final error rates are reported using models trained on the full training set of 50k images.
Table~\ref{tab:cifar10} summarizes results for CIFAR-10 experiments.
For fair comparison we also show data augmentation strategies in the last column of Table~\ref{tab:cifar10}.
Majority of methods use horizontal flip and random crop, while Tang~\etal~\cite{Tang2013} use horizontal flip and color jitter, which may partially explain the surprisingly low error rate (11.9\%) obtained with LeNet.
In most test cases considering the same architecture and data augmentation strategy, our regularizer produces lower error rates than all other competitive methods.
The only exception is EM-Softmax with LiuNet, which achieves 4.98\% error (ours 5.85\%).
However, their result is obtained on an ensemble of 2 LiuNet models, while our results are measured on a single LiuNet model without any ensemble.
Moreover, it can be seen that our method also provides significant absolute improvements to all six reference models with different architectures.

\begin{table}[!t]
 \centering
 \resizebox{1.0\linewidth}{!}{
  \begin{threeparttable}
  \caption{CIFAR-100 Test Error Rates}\label{tab:cifar100}
  \begin{tabular}{ccccc}
   \toprule
   Method                      & Architecture & Error (\%)      & Margin & Augmentation  \\
   \midrule
   DisturbLabel~\cite{Xie2016} & LeNet        & 41.84           & -      & hflip \& crop \\
   CRN~\cite{An2015}           & VGG16        & 34.4            & -      & -             \\
   L-Softmax~\cite{Liu2016}    & LiuNet       & 29.53           & -      & -             \\
   L-Softmax~\cite{Liu2016}    & LiuNet       & 28.04~\tnote{a} & -      & hflip \& crop \\
   EM-Softmax~\cite{Wang2018}  & LiuNet       & 24.04~\tnote{b} & -      & hflip \& crop \\
   \midrule
   Reference                   & LeNet        & 43.30           & 0.11   & hflip \& crop \\
   Ours                        & LeNet        & 41.68           & 0.23   & hflip \& crop \\
   Reference                   & NIN          & 37.75           & 0.12   & hflip \& crop \\
   Ours                        & NIN          & 34.49           & 0.21   & hflip \& crop \\
   Reference                   & LiuNet       & 26.87           & 0.10   & hflip \& crop \\
   Ours                        & LiuNet       & 25.91           & 0.18   & hflip \& crop \\
   Reference                   & ResNet-20    & 33.20           & 0.05   & hflip \& crop \\
   Ours                        & ResNet-20    & 32.96           & 0.16   & hflip \& crop \\
   Reference                   & ResNet-56    & 26.70           & 0.06   & hflip \& crop \\
   Ours                        & ResNet-56    & 26.54           & 0.14   & hflip \& crop \\
   Reference                   & DenseNet-40  & 25.93           & 0.04   & hflip \& crop \\
   Ours                        & DenseNet-40  & 25.62           & 0.11   & hflip \& crop \\
   \bottomrule
  \end{tabular}
  \begin{tablenotes}
   \item[a] We copy the result from a non-official implementation~\cite{Wang2018}, since the original paper~\cite{Liu2016} does not provide such result for CIFAR-100 with data augmentation.
   \item[b] Ensemble of 2 LiuNet models are used.
  \end{tablenotes}
 \end{threeparttable}}
\end{table}

\subsubsection{CIFAR-100}
Similar to the CIFAR-10 experiment, we also evaluate our method on LeNet, NIN, LiuNet, ResNet-20/56, and DenseNet-40 for CIFAR-100.
LeNets, NINs, ResNets and DenseNets are kept the same with the CIFAR-10 experiment except that the output widths of the last fully-connected layers are increased from 10 to 100 for 100-way classification.
For LiuNet, we adopt the CIFAR-100 LiuNet architecture described in~\cite{Liu2016}, which is slightly larger than the CIFAR-10 LiuNet for a fair comparison.
The hyper-parameter tuning and final evaluation protocol are the same with all previous experiments in this paper.
Results are summarized in Table~\ref{tab:cifar100}.
As the original L-Softmax paper~\cite{Liu2016} only provides results without data augmentation on CIFAR-100, we copy the result from~\cite{Wang2018}, as denoted by superscript ``a'' in the table.
It can be seen that our method outperforms DisturbLabel~\cite{Xie2016} and L-Softmax~\cite{Liu2016} under the same architectures.
Again, EM-softmax~\cite{Wang2018} achieves a lower error rate 26.86\% than ours 25.91\% using model ensembling, while we only measure single model performance.
For all six considered architectures, our method is able to provide performance gain to the reference model.

\begin{table}[!t]
 \centering
 \resizebox{1.0\linewidth}{!}{
  \begin{threeparttable}
   \caption{SVHN Test Error Rates}\label{tab:svhn}
   \begin{tabular}{ccccc}
    \toprule
    Method                      & Architecture & Error (\%) & Margin & Augmentation \\
    \midrule
    DisturbLabel~\cite{Xie2016} & LeNet        & 3.27       & -      & crop         \\
    \midrule
    Reference                   & LeNet        & 3.32       & 0.45   & crop         \\
    Ours                        & LeNet        & 3.12       & 0.99   & crop         \\
    Reference                   & NIN          & 2.67       & 0.48   & crop         \\
    Ours                        & NIN          & 2.46       & 1.14   & crop         \\
    Reference                   & LiuNet       & 1.79       & 0.50   & crop         \\
    Ours                        & LiuNet       & 1.61       & 1.24   & crop         \\ 
    Reference                   & ResNet-20    & 1.91       & 0.40   & crop         \\
    Ours                        & ResNet-20    & 1.82       & 1.17   & crop         \\
    Reference                   & ResNet-56    & 1.72       & 0.54   & crop         \\
    Ours                        & ResNet-56    & 1.63       & 1.04   & crop         \\
    Reference                   & DenseNet-40  & 1.79       & 0.46   & crop         \\
    Ours                        & DenseNet-40  & 1.66       & 0.99   & crop         \\
    \bottomrule
   \end{tabular}
  \end{threeparttable}}
\end{table}

\subsubsection{SVHN}
For SVHN, we still validate our method on the same six architectures as in CIFAR-10 experiments.
The protocol for hyper-parameter tuning and final evaluation is also the same.
Since SVHN is a digit classification task where the semantics of a sample is generally not kept if we flip it horizontally, we do not use flip for data augmentation for this dataset. 
Table~\ref{tab:svhn} summarizes the results of our SVHN experiments.
Many of our considered competitive methods do not perform SVHN experiments, hence we do not have their results in the table.
For LeNet, our method achieves lower error (3.12\%) than DisturbLabel (3.27\%).
Compared with reference models, our method is able to provide consistent performance improvement for all six network architectures.

\begin{table}[!t]
 \centering
 \begin{threeparttable}
  \caption{ImageNet top-1 Validation Error Rates}\label{tab:imagenet}
  \begin{tabular}{cccc}
   \toprule
   Method                      & Architecture & Margin & Error (\%) \\
   \midrule
   Reference                   & ResNet-18    & 0.70   & 30.23      \\
   Ours                        & ResNet-18    & 1.33   & 29.94      \\
   Reference                   & ResNet-50    & 0.82   & 23.85      \\
   Ours                        & ResNet-50    & 1.74   & 23.54      \\
   Reference                   & SENet-50     & 1.19   & 22.37      \\
   Ours                        & SENet-50     & 1.92   & 22.19      \\
   \bottomrule
  \end{tabular}
 \end{threeparttable}
\end{table}

\subsubsection{ImageNet}\label{sec:exp_imagenet}
ImageNet is a large-scale image classification benchmark dataset containing millions of high resolution images in 1000 classes.
We test our method on it using three DNN architectures: ResNet-18/50~\cite{He2016} and SENet-50~\cite{Hu2018}.
For efficiency, we collect well-trained models from the community~\footnote{\url{https://github.com/Cadene/pretrained-models.pytorch}}, and fine-tune them with our regularizer.
Results are summarized in Table~\ref{tab:imagenet}.
We see our method provides consistent accuracy gain for all considered architectures, validating the effectiveness of our regularizer on large-scale datasets with modern DNN architectures.

\subsection{Computational Cost}
Since our method invokes iterative updates to approximate the classification margin and it utilizes high-order gradients during optimization, higher computational cost may be inevitable.
In practice, our method usually requires 6-14$\times$ more wall clock time per epoch and 2-4$\times$ GPU memory than the natural cross-entropy training, depending on the network architecture.
Notice that much less epochs are required when fine-tuning a pre-trained model, thus we advocate a two-step training pipeline as introduced in Section~\ref{sec:exp_imagenet} for large-scale problems.

\section{Conclusion}\label{sec:conclusion}
In this paper, we study the generalization ability of DNNs and aim at improving it, by investigating the classification margin in the input data space, and deriving a novel and principled regularizer to enlarge it.
We exploit the DeepFool adversarial perturbation as a proxy for the margin, and incorporate the $\ell_2$ norm-based perturbation into the regularizer.
The proposed regularization can be jointly optimized with the original classification objective, just like training a recursive network.
By developing proper aggregation functions and shrinkage functions, we improve the classification margin in a direct way.
Extensive experiments on MNIST, CIFAR-10/100, SVHN and ImageNet with modern DNN architectures demonstrate the effectiveness of our method.

\section*{Acknowledgments}
This work is funded by the NSFC (Grant No. 61876095), and the Beijing Academy of Artificial Intelligence (BAAI).

\bibliographystyle{IEEEtran}
\bibliography{ref}

\begin{IEEEbiography}[{\includegraphics[width=1in,height=1.25in,clip,keepaspectratio]{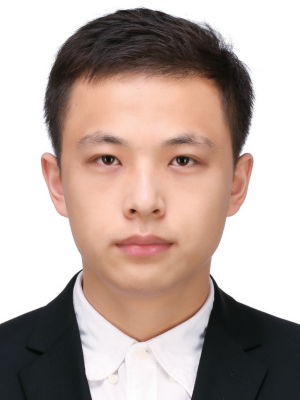}}]{Ziang Yan} received the B.E. degree from Tsinghua University, Beijing, China in 2015. He is currently working toward the Ph.D. degree with the Department of Automation in Tsinghua University, Beijing, China. His current research interests include computer vision and machine learning.
\end{IEEEbiography}

\begin{IEEEbiography}[{\includegraphics[width=1in,height=1.25in,clip,keepaspectratio]{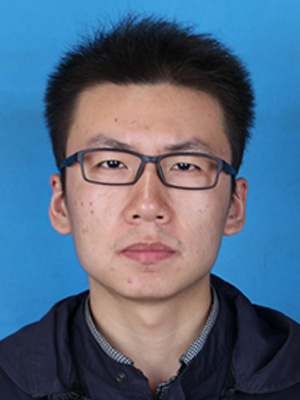}}]{Yiwen Guo} received the B.E. degree from Wuhan University, Wuhan, China, in 2011, and the Ph.D. degree from Tsinghua University, Beijing, China, in 2016. He is with Bytedance AI Lab. Before joining Bytedance in 2019, he was a Staff Research Scientist at Intel Corporation. His current research interests include computer vision and machine learning.
\end{IEEEbiography}

\begin{IEEEbiography}[{\includegraphics[width=1in,height=1.25in,clip,keepaspectratio]{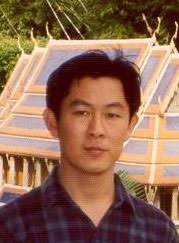}}]{Changshui Zhang} (M'02-SM'15-F'18) received the B.S. degree in mathematics from Peking University, Beijing, China, in 1986, and the M.S. and Ph.D. degrees in control science and engineering from Tsinghua University, Beijing, in 1989 and 1992, respectively. In 1992, he joined the Department of Automation, Tsinghua University, where he is currently a Professor. He has authored more than 200 articles. His current research interests include pattern recognition and machine learning.
\end{IEEEbiography}

\newpage


\section*{Supplementary material}

# Appendices for Adversarial Margin Maximization Networks

Ziang Yan, Yiwen Guo, and Changshui Zhang, *Fellow, IEEE*

---------------◆---------------

TABLE 1
Hyper-parameters for Different Regularizer Configurations

| Aggregation | Shrinkage | $\lambda$ | c | d |
|---|---|---|---|---|
| AVG | LIN | 1 | 4 | 4 |
| AVG | INV | 32 | 4 | 2 |
| AVG | EXP | 32 | 2 | 2 |
| MIN | LIN | 1 | 0.5 | 4 |
| MIN | INV | 32 | 1 | 0.5 |
| MIN | EXP | 32 | 2 | 1 |

TABLE 2
Some Hyper-parameters in the Training process

| Dataset | Architecture | Batch | #Epoch | Init lr | Lr decay @epoch (multiplier) | Scratch |
|---|---|---|---|---|---|---|
| MNIST | MLP | 100 | 100 | 5e-3 | 50 (0.1x), 80 (0.01x) | yes |
| | LeNet | 100 | 100 | 5e-3 | 50 (0.1x), 80 (0.01x) | yes |
| | LiuNet | 100 | 100 | 5e-3 | 50 (0.1x), 80 (0.01x) | no |
| CIFAR-10 | LeNet | 100 | 160 | 5e-3 | 100 (0.1x), 140 (0.01x) | yes |
| | NIN | 100 | 400 | 5e-3 | 250 (0.1x), 350 (0.01x) | no |
| | LiuNet | 100 | 400 | 5e-3 | 250 (0.1x), 350 (0.01x) | no |
| | ResNet-20 | 128 | 400 | 1e-2 | 250 (0.1x), 350 (0.01x) | no |
| | ResNet-56 | 128 | 400 | 1e-2 | 250 (0.1x), 350 (0.01x) | no |
| | DenseNet-40 | 128 | 400 | 1e-2 | 250 (0.1x), 350 (0.01x) | no |
| CIFAR-100 | LeNet | 100 | 160 | 5e-3 | 100 (0.1x), 140 (0.01x) | yes |
| | NIN | 100 | 400 | 5e-3 | 250 (0.1x), 350 (0.01x) | no |
| | LiuNet | 100 | 400 | 5e-3 | 250 (0.1x), 350 (0.01x) | no |
| | ResNet-20 | 128 | 400 | 1e-2 | 250 (0.1x), 350 (0.01x) | no |
| | ResNet-56 | 128 | 400 | 1e-2 | 250 (0.1x), 350 (0.01x) | no |
| | DenseNet-40 | 128 | 400 | 1e-2 | 250 (0.1x), 350 (0.01x) | no |
| SVHN | LeNet | 100 | 100 | 5e-3 | 50 (0.1x), 80 (0.01x) | yes |
| | NIN | 100 | 100 | 5e-3 | 50 (0.1x), 80 (0.01x) | no |
| | LiuNet | 100 | 100 | 5e-3 | 50 (0.1x), 80 (0.01x) | no |
| | ResNet-20 | 128 | 100 | 5e-3 | 50 (0.1x), 80 (0.01x) | no |
| | ResNet-56 | 128 | 100 | 5e-3 | 50 (0.1x), 80 (0.01x) | no |
| | DenseNet-40 | 128 | 100 | 5e-3 | 50 (0.1x), 80 (0.01x) | no |
| ImageNet | ResNet-18 | 32 | 10 | 1e-2 | 5 (0.1x), 8 (0.01x) | no |
| | ResNet-50 | 32 | 10 | 1e-2 | 5 (0.1x), 8 (0.01x) | no |
| | SENet-50 | 32 | 20 | 1e-2 | 10 (0.1x), 15 (0.01x) | no |

## APPENDIX A
## HYPER-PARAMETERS IN THE MNIST EXPLORATORY EXPERIMENTS

In this paper we perform grid search on hyper-parameters $\lambda$, c and d for each combination of aggregation function and shrinkage function on MLP models. Table 1 shows obtained values for these three hyper-parameters.

## APPENDIX B
## SOME HYPER-PARAMETERS FOR TRAINING

Table 2 summarizes some hyper-parameters in the training protocol, including the batch size, number of epochs, initial learning rate and how it decays.

## APPENDIX C
## DISCUSSION FOR [1]

A contemporaneous work known as MMA [1] also attempts to maximize the margin in the input space, for model robustness just like in [2]. Although it shares the similar ambition (i.e., input space margin maximization) to our work, there are important differences:

- It aims to maximize the average margin. As explained and demonstrated in our Section 3.2 and Section 3.3, simply optimizing the average margin may hurt the generalization performance, and the MMA-trained models indeed achieves lower accuracies on clean test images than the naturally trained counterparts (e.g., they report at most 89.4% test accuracy, while the naturally trained model shows ∼95.8%). By contrast, we develop various shrinkage and aggregation functions to fulfill the theoretical requirement, and our experiments also show improved generalization performance on all test cases.

- To calculate the gradient of the margin-related term, MMA utilizes the alignment between margin maximization and adversarial training, while we instead backward through the attacking mechanism.


\end{document}